\documentclass[letterpaper, 10 pt, conference]{ieeeconf}  % Comment this line out if you need a4paper

\IEEEoverridecommandlockouts                              % This command is only needed if 
\usepackage{cite}
\usepackage{graphicx}
\usepackage{times}
\usepackage{epsfig}
\usepackage{graphicx}
\usepackage{amsmath, bm}
\usepackage{amssymb}
\usepackage{comment}
\usepackage{subcaption}
\usepackage{caption}

\usepackage{tabularx}
\usepackage{arydshln}
\usepackage{hhline}

\usepackage{graphicx}
\usepackage{caption}
\usepackage{subcaption}
\usepackage{pdflscape}
\usepackage{cite}

\usepackage{amsmath}
\usepackage{amssymb}
\usepackage{algorithm}
\usepackage{booktabs}
\usepackage[noend]{algpseudocode}

\usepackage[colorlinks=true,citecolor=black,linkcolor=black,urlcolor=black]{hyperref}

\usepackage[subtle,title=normal]{savetrees}

\usepackage[english]{babel}
\usepackage[utf8]{inputenc}

\usepackage[colorinlistoftodos]{todonotes}
\usepackage{hyperref}
\usepackage{cite}
\usepackage{array}
\usepackage[font=footnotesize,labelfont=bf]{caption}

\newcolumntype{L}[1]{>{\raggedright\let\newline\\\arraybackslash\hspace{0pt}}m{#1}}
\newcolumntype{C}[1]{>{\centering\let\newline\\\arraybackslash\hspace{0pt}}m{#1}}
\newcolumntype{R}[1]{>{\raggedleft\let\newline\\\arraybackslash\hspace{0pt}}m{#1}}
\hypersetup{
    colorlinks=true,
    linkcolor=blue,
    filecolor=magenta,      
    urlcolor=cyan,
}
\usepackage{bbm}

\title{\LARGE \bf
Self-Supervised Correspondence in Visuomotor Policy Learning
}

\author{Peter Florence$^{1}$, Lucas Manuelli$^{1}$, and Russ Tedrake$^{1}$% <-this % stops a space
\thanks{Video, source code, at: \href{https://sites.google.com/view/visuomotor-correspondence}{sites.google.com/view/visuomotor-correspondence}}%%
\thanks{$^{1}$All authors are with the Computer Science and Artificial Intelligence Laboratory (CSAIL),
        Massachusetts Institute of Technology, 32 Vassar St., Cambridge, MA, USA
        {\tt\small \{peteflo,manuelli,russt\}@csail.mit.edu}}%
}

\begin{document}

\twocolumn[{%
\renewcommand\twocolumn[1][]{#1}%
%\maketitle

%\vspace{-1.3cm}
\begin{center}
		{
		\vspace{0.5cm}
		{\LARGE \bf
		Self-Supervised Correspondence in Visuomotor Policy Learning}
		
		\vspace{0.5cm}
		\normalsize
		Peter Florence$^{1}$~~~~~~Lucas Manuelli$^{1}$~~~~~~Russ Tedrake$^1$\\ \vspace{0.3cm}}
\centering

\vspace{1.0cm}
%\includegraphics[width=\linewidth]{images/teaser.pdf}
%    \captionof{figure}{DeepSDF represents signed distance functions (SDFs) of shapes via latent code-conditioned feed-forward decoder networks. Above images are raycast renderings of DeepSDF interpolating between two shapes in the learned shape latent space. Best viewed digitally.}    \label{fig:teaser}
\end{center}%
}]

\let\thefootnote\relax\footnotetext{\text{Video, source code, at: \href{https://sites.google.com/view/visuomotor-correspondence}{sites.google.com/view/visuomotor-correspondence}}} 
\let\thefootnote\relax\footnotetext{$^{1}$All authors are with the Computer Science and Artificial Intelligence Laboratory (CSAIL), Massachusetts Institute of Technology, 32 Vassar St., Cambridge, MA, USA. \tt\small \{peteflo,manuelli,russt\}@csail.mit.edu}

%\maketitle 
%%\thispagestyle{empty} 
%%\pagestyle{empty} 

%%%%%%%%%%%%%%%%%%%%%%%%%%%%%%%%%%%%%%%%%%%%%%%%%%%%%%%%%%%%%%%%%%%%%%%%%%%%%%%%
\begin{abstract}

%How can robots best perform self-supervised visual learning in order to aid visuomotor policy learning?  
In this paper we explore using self-supervised correspondence for improving the generalization performance and sample efficiency of visuomotor policy learning.  Prior work has primarily used approaches such as autoencoding, pose-based losses, and end-to-end policy optimization in order to train the visual portion of visuomotor policies. We instead propose an approach using self-supervised dense visual correspondence training, and show this enables visuomotor policy learning with surprisingly high generalization performance with modest amounts of data: using imitation learning, we demonstrate extensive hardware validation on challenging manipulation tasks with as few as 50 demonstrations. Our learned policies can generalize across classes of objects, react to deformable object configurations, and manipulate textureless symmetrical objects in a variety of backgrounds, all with closed-loop, real-time vision-based policies. Simulated imitation learning experiments suggest that correspondence training offers sample complexity and generalization benefits compared to autoencoding and end-to-end training. 
%\pf{Maybe mention: Modularity? As few as 50 demonstrations?}

\end{abstract}

%%%%%%%%%%%%%%%%%%%%%%%%%%%%%%%%%%%%%%%%%%%%%%%%%%%%%%%%%%%%%%%%%%%%%%%%%%%%%%%%
\section{INTRODUCTION}

To achieve general-purpose manipulation skills, robots will need to use vision-based policies and learn new tasks in a scalable fashion with limited human supervision. % A primary challenge is in training effective policies, with limited data, that operate on raw images.  
For visual training, prior work has often used methods such as end-to-end training \cite{levine2016end}, autoencoding \cite{finn2016deep}, and pose-based losses \cite{levine2016end,zhang2018deep}.  These methods, however, have not benefitted from the rich sources of self-supervision that may be provided by dense three-dimensional computer vision techniques \cite{mahjourian2018unsupervised,pillai2019superdepth,schmidt2017self}, for example correspondence learning which robots can automate without human input \cite{florence2018dense}. % \pf{Examples: monodepth, correspondence learning, useful for robots?}  %which for example has underpinned recent advances in monocular depth estimation, robot-supervised correspondence learning \cite{florence2018dense}.

\begin{figure}
\centering
\hspace*{-0.2cm}
  \includegraphics[keepaspectratio=true,scale=0.33]{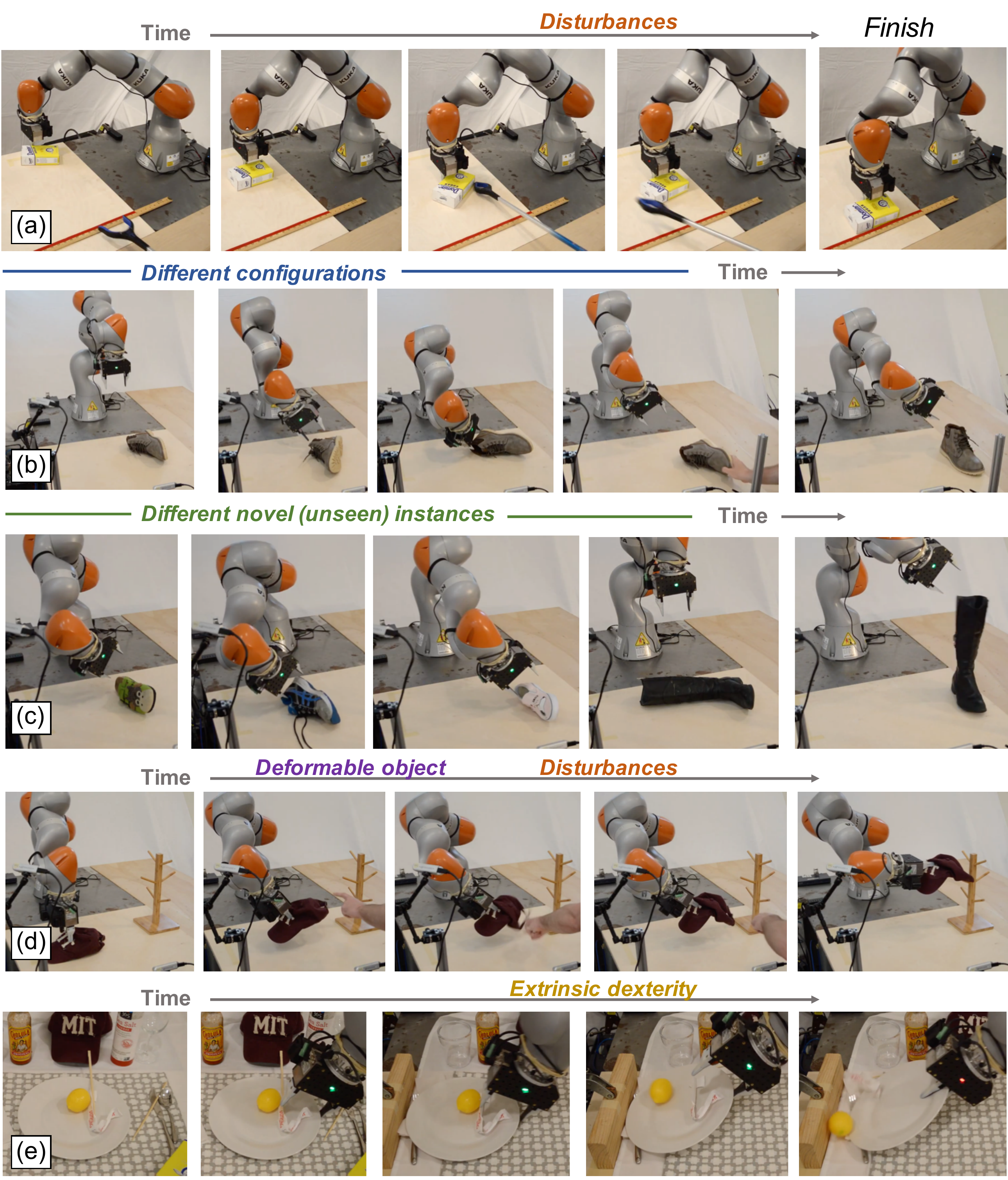}
  \captionof{figure}{Examples of autonomous policies, including a variety of non-prehensile, class-general, and deformable manipulation.
  %(a) pushing, (b,c) flipping, (d) placing deformable objects with disturbances, and (e) pushing, raising, sliding, and grasping. 
  Table \ref{table_hardware_results} details hardware results.}\label{hardwarefig}
% \vspace{-1.0cm} % This helps get everything on Page 1
\end{figure}

Correspondence is fundamental in computer vision, and we believe it has fundamental usefulness for robots learning complex tasks requiring visual feedback.  In this paper we introduce using self-supervised correspondence for visuomotor policies, and our results suggest this enables policy learning that is surprisingly capable. Our evaluations pair correspondence training with a simple imitation learning objective, and
%Using dense self-supervised correspondence training together with imitation learning, 
extensive hardware validation shows that learned policies can address challenging scenarios: manipulating deformable objects, generalizing across a class of objects, and visual challenges  such as textureless objects, clutter, moderate occlusion, and lighting variation (Fig~\ref{hardwarefig}).
%, with as few as 50 demonstrations and no further environment interaction.  
Additionally our simulation-based comparisons empirically suggest that 
our method offers significant generalization and sample complexity advantages compared to existing methods for training visuomotor policies, while requiring no additional human supervision. To bound our method's scope: while spatial correspondence alone cannot suffice for all tasks (for example, it cannot discriminate when to be finished cooking eggs), there is a wide set of tasks for which dense spatial correspondence may be useful: essentially any spatial manipulation task. %essentially any task which comprises manipulating objects spatially.

\textbf{Contributions.} Our primary contribution is (i) a novel formulation of visuomotor policy learning using self-supervised correspondence. 
Through simulation experiments (ii) we measure that this approach offers sample complexity and generalization benefits compared to a variety of baselines, and (iii) we validate our method in real world experiments.
We believe that compared to the existing state of the art in robotic manipulation, the abilities of our learned policies represent exciting levels of performance, especially the generalization across challenging scenarios (category-level manipulation, deformable objects, visually challenging scenes) and with limited data (between 50 and 150 demonstrations).
We also (iv) introduce a novel data augmentation technique for behavior cloning, and (v) demonstrate a new technique for multi-camera time-synchronized dense spatial correspondence learning.

\section{RELATED WORK}

%There are a number of primary related areas of work to this paper.  
We focus our related work review around two topics: visual training methods for visuomotor policies, and approaches for providing the policy learning signal. An overview of the broader topic of robot learning in manipulation is provided in \cite{kroemer2019review}.  For more related work in self-supervised robotic visual learning, including correspondence learning, we refer to the reader to \cite{florence2018dense}.

\subsection{Visual Training Methods for Visuomotor Policies}\label{vis-training-methods}

There have been three primary methods used in the robot learning literature to train the visual portion of visuomotor policies.  Often these methods are used together -- for example \cite{levine2016end,zhang2018deep} use pose-based losses together with end-to-end training. 
\textbf{(1) End-to-End training.}  This approach can be applied to any learning signal that is formed as a consequence of a robot's actions, for example through imitation learning or reinforcement learning. While often end-to-end training is complemented with other learning signals, other works use purely end-to-end training.
\textbf{(2) Autoencoders.}  Autoencoding can be applied to any data with no supervision and is commonly used to aid visuomotor policy learning \cite{finn2016deep,yang2016repeatable,finn2016guided,ghadirzadeh2017deep,van2016stable,rahmatizadeh2018vision}. Sometimes  polices are learned with a frozen encoder \cite{finn2016deep,yang2016repeatable,finn2016guided}, other times in conjunction with end-to-end training \cite{rahmatizadeh2018vision}.
\textbf{(3) Pose-based losses.}  In \cite{levine2016end}, for example, % as a pre-training step 
a separate dataset is collected of the robot holding objects, and assuming that the objects are rigid and graspable, then using the robot's encoders and forward kinematics the visual model can be trained to predict the object pose.
In \cite{zhang2018deep}, pose-based auxiliary losses are used regardless of whether or not objects are held -- we wouldn't expect this to learn how to predict object configurations unless they are also rigid and grasped. 
Simulation-based works \cite{james2017transferring} have also used auxiliary losses for object and gripper positions.

In our comparison experimentation, we include end-to-end training and autoencoding, but not pose-based losses, since they are not applicable to deformable or un-graspable objects.  While the above are three of the most popular, other visual training methods include: training observation dynamics models \cite{agrawal2016learning,ebert2018visual}, using time-contrastive learning \cite{sermanet2018time}, or using no visual training and instead using only generic pre-trained visual features \cite{sermanet2016unsupervised}.
Relevant concurrent works include \cite{kulkarni2019unsupervised} which proposes autoencoder-style visual training but with a reference image and novel architecture, and \cite{sieb2019graph} which proposes a graph-based reward function using a fixed set of correspondences.

\subsection{Methods for Learning Vision-Based Closed-Loop Policies}

%To acquire policies, we must choose a policy learning style, in addition to the visual training methods discussed in the previous section.
While the previous section discussed visual training methods, to acquire policies they must be paired with a policy learning signal. %, for which there exist a wide variety of options in the literature. %this includes: real-world reinforcement learning (cite), imitation learning from human demonstrations (cite), and sim-to-real transfer (cite). 
We are particularly interested in approaches that can (i) scalably address a wide variety of tasks with potentially deformable and unknown objects, (ii) use a small incremental amount of human effort (on the order of 1 human-hour) per each new object or task, and (iii) produce real-time vision-based closed-loop policies.

One source of policy learning signal may be from reinforcement learning, which
has demonstrated many compelling results.  A primary challenge, however,
is the difficulty of measuring rewards in the real world. Some tasks such as grasping can be self-supervised \cite{pinto2016supersizing}, and other tasks can leverage assumptions that objects are grasped and rigid in order to compute rewards \cite{levine2016end}, but this only applies to a subset of tasks. A more generalizable direction may be offered by unsupervised methods of obtaining reward signals \cite{finn2016deep,sermanet2016unsupervised,sermanet2018time}. %se instrumented environments for reward signals \cite{yahya2017collective}.  
Another direction which has shown promising results is using sim-to-real transfer \cite{james2017transferring,zhu2018reinforcement,matas2018sim,andrychowicz2018learning}, but our interest in a small amount of incremental human effort per new task is challenging for these methods, since they currently require significant engineering effort for each new simulation scenario.

Another powerful source of signal may come from imitation learning from demonstration, which several recent works have shown promise in using to produce real-time vision-based closed-loop policies \cite{yang2016repeatable,finn2016guided,sermanet2016unsupervised,finn2017one,rahmatizadeh2018vision,zhang2018deep,yu2018one,sermanet2018time}. 
We point the reader to a number of existing reviews of learning from demonstration \cite{argall2009survey,billard2008robot}.  %We could not find any criteria-satisfying works whose policies use common robotics perception concepts such as object poses (which don't apply to deformable objects), or any other explicit object/environment representation. 
Another direction may be to learn models from observations and specify goals via observations \cite{finn2016deep,agrawal2016learning,ebert2018visual}, but these may be limited to tasks for which autonomous exploration has a reasonable chance of success. 
In terms of limitations of these prior works, one primary challenge relates to reliability and sample complexity 
-- it is not clear how much data and training would be required in order to achieve any given level of reliability.  Relatedly, a second limitation is that little work has characterized the distributions over which these methods should be trained and subsequently expected to generalize.  Third, like in many areas of robotics it is difficult to reproduce results and compare approaches on a common set of metrics.  %This is much unlike in the computer vision community, where standard datasets are used and the software to reproduce results is often made available.  
%Instead, to attempt to reproduce hardware results requires expensive robots, significant engineering efforts, and never can all of the physical variation of real-world manipulation tasks be perfectly replicated.  
While we believe hardware validation is critical, we also believe that increased effort should be put into simulation-based results that compare methods and can be reproduced. 

\section{VISUOMOTOR FORMULATION}
\label{sec:formulation}

First as preliminary we identify some primary attributes of existing approaches in visuomotor policy learning (Sec.~\ref{prelim-vis-motor}). We then present our approach based on self-supervised correspondence (Sec.~\ref{subsec:correspondence-visuomotor}). The discussion of visuomotor policy learning in this section is agnostic to the specific learning algorithm, i.e. reinforcement learning, imitation learning, etc., and focuses %on the structure of the visuomotor-policy 
on the model structure and sets of trainable parameters. Sec.~\ref{sec:methodology} discusses the application of our approach to a specific case of imitation learning.

\subsection{Preliminary: Visuomotor Policies} \label{prelim-vis-motor}

\begin{figure*}
\centering
\hspace*{-0.2cm}
  \includegraphics[keepaspectratio=true,scale=0.33]{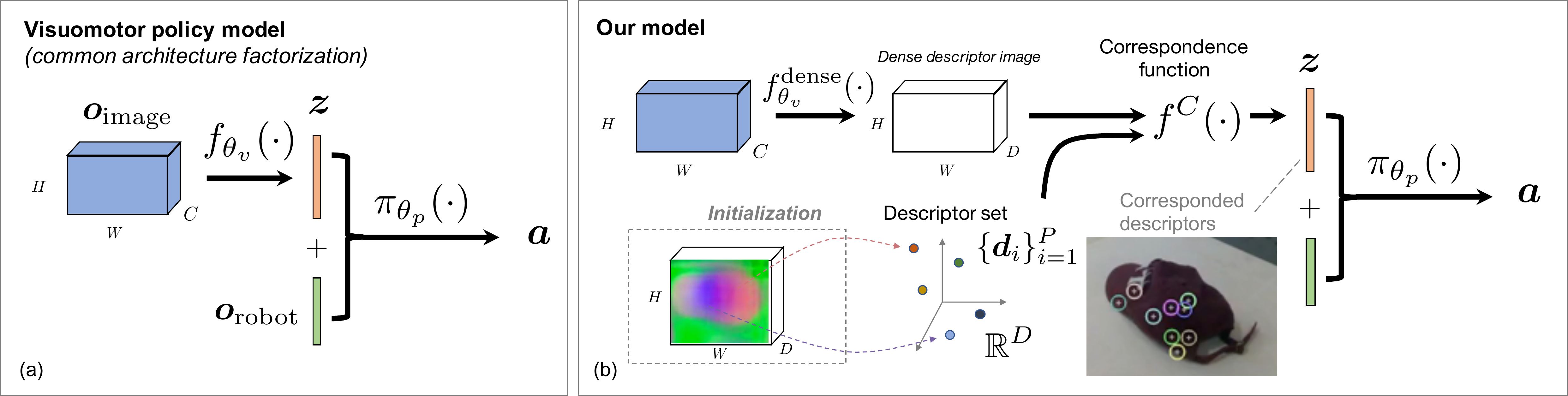}
  \captionof{figure}{Diagram of common visuomotor policy factorization (a), and our proposed model (b) using visual models trained on correspondence.}\label{visuomotor_diagrams}
% \vspace{-0.5cm} % This helps get everything on Page 1
\end{figure*}

We would like to have 
a policy $\bm{a}_t = \pi_\theta(\bm{o}_{0:t})$, where $\bm{o}_{0:t} = (\bm{o}_0, \bm{o}_1, \ldots \bm{o}_t)$ is the full sequence of the robot's observations during some episode up until time $t$, with each $\bm{o}_i \in \mathcal{O}$, the robot's observation space. This sequence of observations is mapped by $\pi_\theta(\cdot)$, the robot's policy parameterized by $\theta$, to the robot's actions $\bm{a}$, each $\in \mathcal{A}$.  In particular, we are interested in visuomotor policies in which the observation space contains high-dimensional images $\mathcal{O}_{\text{image}} \subset \mathcal{O}$, for example $\mathcal{O}_{\text{image}} = \mathbb{R}^{W \times H \times C}$ for a $C$-channel, width $W$, and height $H$ image.  The visual data is perhaps complemented with additional lower-dimensional measurements $\mathcal{O}_{\text{robot}}$, such as produced from sensors like the robot's encoders, such that $\mathcal{O}_{\text{robot}} \times \mathcal{O}_{\text{image}} = \mathcal{O}$.

It is common for a visuomotor policy to have an architecture that can factored as displayed in Fig.~\ref{visuomotor_diagrams}(a),
\begin{align}
     & \bm{z} = f_{\theta_v}(\bm{o}_{\text{image}})         :  \bm{o}_{\text{image}} \in \mathcal{O}_{\text{image}}, \  \bm{z} \in \mathbb{R}^Z \\ 
     & \bm{a} = \pi_{\theta_p}(\bm{z}, \bm{o}_{\text{robot}} ) :   \bm{o}_{\text{robot}} \in \mathcal{O}_{\text{robot}}, \ \bm{z} \in \mathbb{R}^Z, \  \bm{a} \in \mathcal{A}
\end{align}
in which a visual model $f_{\theta_v}(\cdot)$, parameterized by $\theta_v$, processes the high-dimensional $\bm{o}_{\text{image}}$ into a much smaller $Z$-dimensional representation $\bm{z}$. The policy model $\pi_{\theta_p}(\cdot)$ then combines the output of the visual model with other observations $\bm{o}_{\text{robot}}$.  This is a practical modeling choice -- images are extremely high dimensional, i.e. in this work we use images in $\mathbb{R}^{640 \times 480 \times 3} = \mathbb{R}^{921,600}$, whereas our $\mathcal{O}_{\text{robot}}$ is at most $\mathbb{R}^{13}$.  
A wide variety of works have employed a similar architecture to \cite{levine2016end}, consisting of convolutional networks extracting features from raw images into an approximately $Z=32$ to $100$ bottleneck representation of features, e.g. %\cite{levine2016end,finn2016deep,zhang2018deep}.  
\cite{finn2016deep,rahmatizadeh2018vision,finn2017one,yahya2017collective,ghadirzadeh2017deep,zhang2018deep}

\subsection{Visual Correspondence Models for Visuomotor Policy Learning} \label{subsec:correspondence-visuomotor}
The objective of the visual model is to produce a feature vector $\bm{z}$ which serves as a suitable input for policy learning. %In our simulation benchmarks we have access to the ground truth state of objects and so can encode this information directly in $\bm{z}$.
In particular, we are interested in deploying policies that can operate directly on RGB images.
Given the role that pose estimation has played in traditional manipulation pipelines it seems valuable to encode the configuration of objects of interest in the vector $\bm{z}$. Pose estimation, however, doesn't extend to the cases of deformable or unknown objects. %One approach would be to incorporate a pose estimator which estimates the pose of the object(s) of interest from the RGB image. However this approach doesn't extend to the case of deformable or unknown objects. 
Some of the prior works discussed in Sec.~\ref{vis-training-methods}, for example \cite{levine2016end,finn2016deep}, have interpreted their learned feature points $\bm{z}$  as encoding useful spatial information for the objects and task. These feature points are learned via the supervisory signals of end-to-end, pose-based, or autoencoding losses, and don't explicitly train for spatial correspondence.
In contrast our approach is to directly employ visual correspondence training,
%These approaches use indirect sources of supervision, such as autoencoding or end-to-end training (outlined in Section \ref{vis-training-methods}), which don't explicitly train for the task of visual correspondence. 
building off the approach of 
\cite{florence2018dense}  which can in a self-supervised manner, learn pixel descriptors of objects that are effective in finding correspondences between RGB images.
% Rather than indirectly training for $\bm{z}$ encoding spatial information, we can leverage the approach in
%Instead of using these indirect methods of visual supervision we build off the approach of 

We introduce four different methods for how to employ dense correspondence models as the visual basis of visuomotor policy learning. The first three are based on the idea of a set of points on the object(s) that are localized either in image-space or 3D space. %. %These points often lie on the object(s) being manipulated and provide information about object configuration. 
We represent these points as a set $\{\bm{d}_i\}_{i=1}^P$ of $P$ descriptors, with each $\bm{d}_i \in \mathbb{R}^D$ representing some vector in the $D$-dimensional descriptor space produced by a dense descriptor model $f^{\text{dense}}_{\theta_v}(\cdot)$. This  $f^{\text{dense}}_{\theta_v}(\cdot)$, a deep CNN, maps a full-resolution RGB image, $\mathbb{R}^{W \times H \times 3}$, to a full-resolution descriptor image, $\mathbb{R}^{W \times H \times D}$. Let us term $f^{C}(\cdot)$ to be the non-parametric correspondence function that, given one or more descriptors and a dense descriptor image $f^{\text{dense}}_{\theta_v}(\bm{o}_{\text{image}})$, provides the predicted location of the descriptor(s): 
\begin{align}
\label{eq:z_dc}
%\underset{ \Theta } {min } \ \  & \mathcal{L}  \bigg( \pi_{\theta_p} \big(\bm{z}, \bm{o}_{\text{robot}}    \big) \bigg)   \\ % (f_{c}(f_{dense},% $\{d_i\}_{i=1}^P$)), o_{other} )\\ 
& \bm{z} = f^{C} \big( f^{\text{dense}}_{\theta_v}(\bm{o}_{\text{image}}),  \{\bm{d}_i\}_{i=1}^P \big)
\end{align}
Specifically $f^C:\mathbb{R}^{W \times H \times D} \times \mathbb{R}^{P \times D} \rightarrow \mathbb{R}^{P \times K}$, where $K = 2$ corresponds to: $\bm{z}$ is the predicted corresponding $(u,v)$ pixel coordinates of each descriptor in the image, %while $K=2$, %if $\bm{z} = f^C(f^{dense}_{\theta_v}(\bm{o}_{\text{image}}))$ 
%$\bm{z}$ contains the $(u,v)$ pixel location in the RGB image for each point $\bm{d}_i$, 
while $K=3$ is their predicted 3D coordinates.\footnote{The specific form of $f^{C}(\cdot)$ is defined by how the correspondence model was trained. In our preferred model we compute a spatial-expectation using a correspondence kernel, either in image-space or 3D. See \cite{florence2019thesis}, Chapter 4, for details.} All four methods optimize a generic policy-based loss function, shown in Eq. (\ref{eq:generic_loss}), and vary only in the set of learnable parameters $\Theta$ and how $\bm{z}$ is acquired (the first three use Eq.~\ref{eq:z_dc}). %As mentioned at the start of Section \ref{sec:formulation} 
This loss function $\mathcal{L}$ 
is generic and could represent any approach for learning the parameters of a visuomotor policy.% i.e. imitation learning, reinforcement learning, etc.
\begin{align}
\label{eq:generic_loss}
\underset{ \Theta } {min } \ \  & \mathcal{L}  \bigg( \pi_{\theta_p} \big(\bm{z}, \bm{o}_{\text{robot}}    \big) \bigg)   %\\  (f_{c}(f_{dense},% $\{d_i\}_{i=1}^P$)), o_{other} )\\ 
%& \bm{z} = f^{C} \big( f^{dense}_{\theta_v}(\bm{o}_{\text{image}}),  \{\bm{d}_i\}_{i=1}^P \big)
\end{align}
%
%
%\centering
%\footnotesize
%
%

\textbf{Fixed Descriptor Set.} This method only optimizes the policy parameters,  $\Theta  = \{\theta_p\}$. In this case both the set of descriptors $\{\bm{d}_i\}_{i=1}^P$ and visual model $f^{\text{dense}}_{\theta_v}(\cdot)$ are fixed. We use a simple initialization scheme of sampling $\{\bm{d}_i\}$ from a single masked reference descriptor image. While we have found this method to be surprisingly effective, it is unsatisfying that the visual model's representation is not optimized after the random %descriptor set 
initialization process. 

\textbf{Descriptor Set Optimization.} This method optimizes the descriptor set  $\{\bm{d}_i\}_{i=1}^P$ along with the policy parameters $\theta_p$ while keeping the dense descriptor mapping $f^{dense}_{\theta_v}$ fixed. %This is an attractive option that has become our preferred method, both philosophically and also because it provides the best performance.
Intuitively $f^{dense}_{\theta_v}$ has already been trained to perform correspondence, and we are simply allowing the policy optimization to choose \textit{what to correspond}. We have observed that Descriptor Set Optimization can improve validation error in some cases over a Fixed Descriptor Set, and adds minimal computational cost and parameters.
% Specifically, the number of additional parameters will be $P \times D$, and for example we often use $P$ in the range of 16 and $D$ in the range of 10. Interestingly, we observe that the advantage of Descriptor Set Optimization over Fixed Descriptor Sets is most pronounced when $P$ is small and $D$ is large, i.e., a small number of descriptors are used, so it especially matters which descriptors are used, and also we hypothesize that the larger-dimensional descriptors are less prone to local minima than smaller-dimensional descriptors.

\textbf{End-to-End Dense Optimization.} The third option is to train the full model architecture end-to-end by including $\theta_v$ in the optimization. While we may have expected this approach to allow the visual model to more precisely focus its modeling ability on task-critical parts of images, we so far have not observed a performance advantage of this approach over Descriptor Set Optimization.%, and we have observed that policy optimization steps are approximately 50 times slower when optimizing our deep (34-layer ResNet) visual models during policy optimization.

\textbf{End-to-End with Correspondence Pretraining.} The fourth option is to directly apply
a differentiable operation to a model which was previously trained on dense correspondence. We can apply any differentiable operation $g(\cdot)$ on top of $f^{\text{dense}}_{\theta_v}$ directly to produce a representation $\bm{z} =  g \big( f^{dense}_{\theta_v} (\bm{o}_{\text{image}}) \big)$. For example, we can apply non-parametric channel-wise spatial expectations to each of the $D$ channels of the dense descriptor images. The optimization variables in this case are $\Theta = \{\theta_p, \theta_v \}$.

% \begin{center}
%   %\centering
%   \footnotesize
% \begin{tabular}{ |c|c|} 
%  \hline 
%  \textbf{Method} & $\Theta$ \\ 
% \hhline{|=|=|}
% %\hline & & & \hline
%  Fixed Descriptor Set & $\{\theta_p \}$ \\ 
%  \hline
%  Descriptor Set Optimization & $\{\theta_p,\{\bm{d}_i\}_{i=1}^P \}$\\ 
%  \hline
%  End-to-End Dense Optimization &  $\{\theta_p,\{\bm{d}_i\}_{i=1}^P, \theta_v \}$\\ 
%  \hline
%  End-to-End with  & $\{\theta_p, \theta_v \}$ \\ 
%  Correspondence Pretraining & $\bm{z} =  g \big( f^{\text{dense}}_{\theta_v} (\bm{o}_{\text{image}}) \big)$ \\ \hline
% \end{tabular}
%   %\caption{Summary of task attempts and success rates for hardware validation experiments.  Autonomous re-tries are counted as successes.} 
%   \label{optimization_options}
% \end{center}

For our $f^{\text{dense}}_{\theta_v}$ we use a 34-layer ResNet, as in \cite{florence2018dense}, which is a powerful vision backbone. Accordingly, using either a fixed- or optimized- descriptor set will significantly increase policy training speed, since it does not require forward-backward optimizing through a very deep convolutional network in each step of policy training, which in our case is 1 to 2 orders of magnitude faster.

\section{VISUAL IMITATION FORMULATION}
\label{sec:methodology}

We now propose how to use the general approach of Sec.~\ref{subsec:correspondence-visuomotor} for a specific type of imitation learning for robot manipulation.

\subsection{Robot Observation and Action Spaces}
\label{subsec:robot_observation_space}
At the lowest level our controller sends joint velocity commands to the robot. For ease of providing demonstrations via teleoperation, the operator commands relative-to-current desired end-effector poses $T_{\Delta,\text{cmd}}$. A low-level Jacobian based controller then tracks these end-effector pose setpoints. Our learned policies also output $T_{\Delta, \text{cmd}}$. % which we transform to world frame, $T_{\text{cmd}}$, with the known current pose. 
The teleoperator also commands a gripper width setpoint which again is tracked by a low-level controller. Thus the action space is $\bm{a} = (T_{\Delta,\text{cmd}}, w_\text{gripper}) \in \mathcal{A} = SE(3) \times \mathbb{R}^+$. 

Our $\bm{o}_{\text{robot}} \in \mathbb{R}^{13}$ is (i) three 3D points on the hand as in \cite{levine2016end}, (ii) an axis-angle rotation relative to the task's starting pose, and (iii) the gripper width.  As noted previously, $\bm{o}_{\text{image}} \in \mathbb{R}^{921,600}$.

\subsection{Imitation Learning Visuomotor Policies}
\label{subsec:imitation_learning_visuomotor_policies}
To evaluate visual learning strategies for enabling visuomotor policy learning, we use imitation learning via a simple behavioral cloning \cite{pomerleau1989alvinn} strategy, which a few recent works have demonstrated to be viable for learning visuomotor manipulation policies \cite{rahmatizadeh2018vision,zhang2018deep}.  Optimizing a policy with parameters $\Theta$ on the behavioral cloning objective, given a dataset of $N_{\text{train}}$ trajectories of observation-sequence-to-action pairs $\{ (\bm{o}_{t}, \bm{a}^*_t) \}_{t=0}^{T_i}$ can be written as:
\begin{equation}
\underset{\Theta}{min} \frac{1}{N_{\text{train}}} \sum_{i=1}^{N_{\text{train}}} \sum_{t=0}^{T_i} \mathcal{L}_\text{BC} \big( \bm{a}^*_t, \pi_{\Theta}( \bm{o}_{t} ) \big)    
\end{equation}
For our loss function we use a simple weighted sum of $l_1$ and $l_2$ loss, 
$\mathcal{L}_\text{BC}(\cdot) = || \bm{a}^* - \pi(\cdot) ||_2^2 + \lambda || \bm{a}^* - \pi(\cdot) ||_1$
where we use $\lambda = 0.1$. %although many different loss functions could be used. 
We scale $\bm{a}^*$ to equalize 1.0m end-effector translation, 0.1 radians end-effector rotation, and 1.0m gripper translation. 
%One must also choose weightings between translation, rotation, and the gripper (see Appendix).

\subsection{Training for Feedback through Data Augmentation}
\label{subsec:trainig_for_feedback_through_data_augmentation}

We introduce a simple technique which we have found to be effective in at least partially addressing a primary issue in imitation learning: the issue of cascading errors \cite{ross2011reduction}. While other works have shown that injecting noise into the dynamics either during imitation learning \cite{laskey2017dart} or sim-to-real transfer \cite{peng2018sim} can alleviate cascading errors, we provide a simple method based only on data augmentation. This method does not address recovering from discrete changes in the environment, but can address local feedback stabilization.
%but we have observed it does effectively enable the robot to stably execute a trajectory from only a single demonstration, whereas without noise augmentation fitting a policy to a single trajectory does not yield a stabilizing controller. 

Consider the output of our policy in a global frame, $\bm{a} = (T_{\text{cmd}}, w_{\text{gripper}})$, which we can acquire from $T_{\Delta, \text{cmd}}$ since we know the end-effector pose.  %The output of our policy is desired end-effector setpoints and potentially also the gripper width. 
As previously mentioned a low-level controller tracks these setpoints, thus our learned policies can stabilize a trajectory by commanding the same global-frame setpoint $\bm{a}$ in the face of small disturbances to the robot state. %Suppose that $(\bm{o}_t, \bm{a}_t)$ is a training example from a demonstration. Let $\bm{o}_{\text{robot}, t}$ denote the robot-state portion of the observation space. 
If we want our policy to command the same setpoint in the face of a slightly perturbed robot state $\tilde{\bm{o}}_{\text{robot}}$ % = \bm{o}_{\text{robot}} + \eta$ where $\eta \sim \mathcal{N}$ 
we can simply use $((\bm{o}_\text{image}, \tilde{\bm{o}}_{\text{robot}}), \bm{a})$ as an observation-action pair. These noise-augmented observation-action pairs are generated on-the-fly during training. A remaining question, of course, is what scale of noise is appropriate.  In practice given our robot's scale and typical speeds we find $\tilde{\bm{o}}_{\text{robot}} \sim \mathcal{N} (\bm{o}_{\text{robot}}, I\bm{\sigma})$ with $\sigma_i$ of 1mm, 1 degree, and 1cm works well respectively for translational, rotational, and gripper components.

\subsection{Multi-View Time-Synchronized Correspondence Training}
\label{subsec:multi_view_time_synchronized_correspondence_training}

Unlike in previous work which trained robotic-supervised correspondence models only for static environments \cite{florence2018dense}, we now would like to train correspondence models with dynamic environments.  %To address this challenge, we employ a multi-view camera setup with time synchronization. 
Other prior work \cite{schmidt2017self} has used dynamic non-rigid reconstruction \cite{newcombe2015dynamicfusion} to address dynamic scenes. The approach we demonstrate here instead is to correspond pixels between two camera views with images that are approximately synchronized in time, similar to the full-image-embedding training in \cite{sermanet2018time}, but here for pixel-to-pixel correspondence.  

 % into a common frame.  
For training, like the static-scene case, finding pixel correspondences between images requires only %a simple geometric operation involving only
depth images, camera poses, and camera intrinsics.
% For this we require the cameras to be calibrated both for their intrinsics and relative extrinsics.
Autonomous object masking can, similar to \cite{florence2018dense}, be performed using 3D-based background subtraction, using only the live depth sensors' point clouds.  Since both (a) the time-synchronized technique can only correspond between time-synchronized images rather than many different static-scene views (\cite{florence2018dense} used approximately 400]), and (b) the time-synchronized technique does not have access to highly accurate many-view-fused 3D geometry as used in \cite{florence2018dense}, it was unclear that our time-synchronized training would provide as compelling results as shown in Sec.~\ref{subsec:hardware_results}.  To encourage generalization despite having using only two static views, we add rotation, scale, and shear image augmentations, and to help alleviate incorrect correspondences due to noisy depth images, we add photometric-error-based rejection of correspondences.

\subsection{Policy Models}
\label{subsec:policy_models}
We use two standard classes of policy models, Multi-Layer Perceptrons (MLP) and Long Short-Term Memory (LSTM) recurrent networks, which are familiar model classes to many different types of machine learning problems and in particular have been demonstrated to be viable for real-world visuomotor control \cite{levine2016end,zhang2018deep,rahmatizadeh2018vision}. In our evaluations the MLPs are only provided current observations, $\pi(\bm{o}_t)$, whereas through recurrence the LSTMs use the full observation sequence.    %Although these are the models we use, our view is that we are agnostic to which model class is used, as long as it performs well on our metrics.
The Appendix provides more model details.
%For details on the model architectures and training, see the Appendix.

\section{RESULTS}

Our experimentation sought to answer these primary questions: (1) Is it possible to use self-supervised descriptors as successful input to learned visuomotor policies? (2) How does visual correspondence learning compare to the benchmarked methods in terms of enabling effective policy learning, as measured by generalization performance and sample complexity? We also evaluate (3) the effect of noise augmentation, and (4) whether our dynamic-scene visual training technique is capable of effective correspondence learning. Simulation setup and results are detailed in Sec.~\ref{subsec:sim_experiments_and_benchmarking},~\ref{subsec:sim_results}, hardware setup and results are in Sec.~\ref{subsec:real_hardware_experimentation},~\ref{subsec:hardware_results}.

\subsection{Simulation Experimental Setup}
\label{subsec:sim_experiments_and_benchmarking}

\begin{figure}
\centering
\hspace*{-0.4cm}  
  \includegraphics[keepaspectratio=true,width=0.47\textwidth]{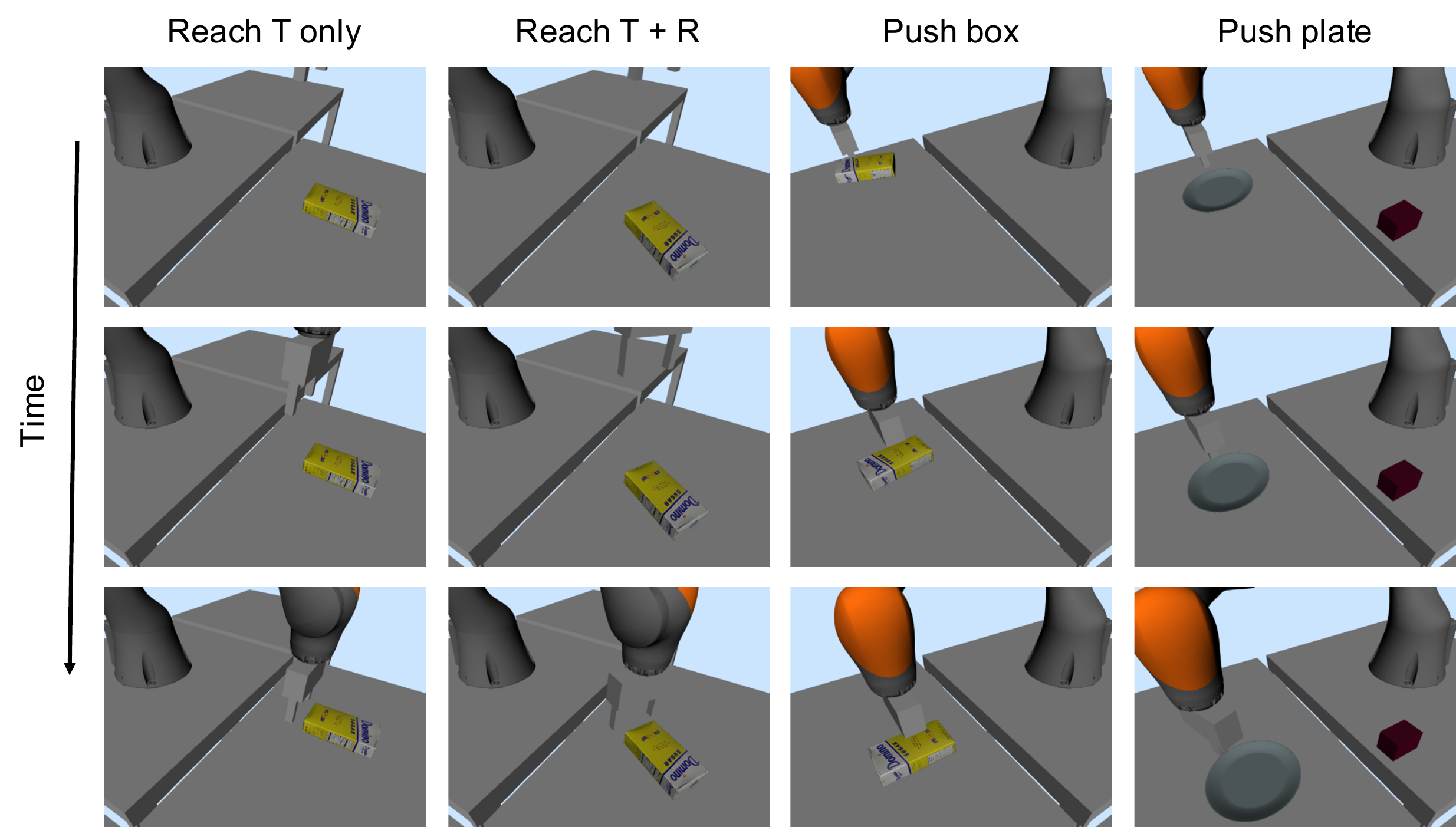}
  \captionof{figure}{RGB images used for visuomotor control in each of the simulation tasks. T=translation, R=rotation, see Appendix for task descriptions. }\label{fig:sim_experiments_images}
\end{figure}

%To compare methods we perform simulated imitation learning experiments (Fig.~\ref{fig:sim_experiments_images}).
We use simulated imitation learning tasks (Fig.~\ref{fig:sim_experiments_images}) to compare the generalization performance of behavior-cloned policies where the only difference is how the ``visual representation'' $\bm{z}$ is acquired. 
%We perform simulated imitation learning experiments on the four tasks shown in Figure \ref{fig:sim_experiments_images}.  
The first two tasks involve reaching to an object whose configuration varies between trials either in translation only, or rotation as well. The additional two tasks are both pushing tasks, which require feedback due to simulated external disturbances. Expert demonstrations use simple hand-designed policies using ground truth object state information.  %Our used 
%
% Similar to our hardware configuration, we use two cameras mounted at a downwards angle towards the robot's table. 
%simulated tasks are:
%
%In all tasks both cameras were used for correspondence training, but we only use one RGB camera for policy input. 
The compared methods are:
%
%The compared methods are:

\begin{enumerate}
\item \textit{Ground truth 3D points (GT-3D):}  $\bm{z}$ is ground truth world-frame 3D locations of points on the object.
\item \textit{Ground truth 2D image coordinates (GT-2D):} $\bm{z}$ is similar to the previous baseline, but the points are projected into the camera using the ground truth camera parameters.% $K$ and extrinsics $^W\bm{T}_{cam}$.
\item \textit{Autoencoder (AE):} $\bm{z}$ is the encoding of a pre-trained autoencoder, similar to the visual training in \cite{finn2016deep,ghadirzadeh2017deep}.  
\item \textit{End-to-End (E2E):}  $\bm{z}$ is the intermediate representation from end-to-end training. This closely resembles the visual training and models in \cite{levine2016end,zhang2018deep}, but we do not also add pose-based losses, in order to investigate end-to-end learning without these auxiliary losses.
\item \textit{Ours, Dense descriptors (DD):}  $\bm{z}$ is the expected image-space locations (DD-2D) or 3D-space locations (DD-3D) of the descriptor set $ \{ \bm{d} \} _i $, where the visual model was trained on dense correspondence. 

\end{enumerate}
Note that the two vision-based baselines AE and E2E share an identical model architecture for producing $\bm{z}$, and differ only in the method used to train the parameters.  The model is close to \cite{levine2016end,finn2016deep,zhang2018deep} with the key architectural traits of having a few convolutional layers followed by a channel-wise spatial expectation operation, which has been widely used %for training visuomotor policies %
\cite{finn2017one,finn2016guided, chebotar2017path,yahya2017collective,yu2019unsupervised,singhiccv17,ghadirzadeh2017deep}. 
Most methods we compare (AE, E2E, DD-2D) use only one RGB camera stream as input to learned policies; DD-3D additionally uses the depth image. DD methods use descriptor set optimization (Sec.~\ref{subsec:correspondence-visuomotor}) and use both views for the correspondence training, before policy training.

See the Appendix for additional model and task details.

\subsection{Simulation Results}\label{subsec:sim_results}

% NEW TABLE
\begin{table}
  \centering
  \footnotesize
  \begin{tabular}{|l|r|r|r|r|}
    \hline
    \multicolumn{1}{|l|}{} & \multicolumn{1}{l|}{Reach} & \multicolumn{1}{l|}{Reach} & \multicolumn{1}{l|}{Push} & \multicolumn{1}{l|}{Push}  \\
    
    \multicolumn{1}{|l|}{Method / Task} & \multicolumn{1}{l|}{T only} & \multicolumn{1}{l|}{T + R} & \multicolumn{1}{l|}{box} & \multicolumn{1}{l|}{plate}  \\ \hline \hline %\hhline{|=|=|=|=|=|}
    
    \textit{Ground truth} 3D points        & 100.0     & 100.0      & 100.0 & 90.5 \\ \hline
    \textit{Ground truth} 2D image coord.  & 94.1      & 95.6       & 100.0 & 92.0 \\ \hline \hline %\hhline{|=|=|=|=|=|}
    \multicolumn{5}{|l|}{\textit{RGB policy input}}  \\ \hline
    Autoencoder, frozen                   &  8.1      & 61.1       & 31.0  & 53.0 \\ \hline
    Autoencoder w/ mask, frozen           & 16.3      & 10.0       & 73.0  & 67.0 \\ \hline
    Autoencoder, then End-to-End          & 40.7      & 38.9       & --  & 16.0 \\ \hline
    End-to-End                             & 43.0      & 32.2       & \bf{100.0} & 5.5 \\ \hline
    End-to-End (34-layer ResNet)           &  --       &  3.3       & --  & -- \\ \hline
    DD 2D image coord. (ours)              & \bf{94.1} & \bf{97.8}  & \bf{100.0} & \bf{87.0} \\ \hline \hline %\hhline{|=|=|=|=|=|}
    \multicolumn{5}{|l|}{\textit{RGBD policy input}}  \\ \hline
    DD 3D coord. (ours)                    & \bf{100.0}  & \bf{100.0}    & --  & \bf{98.0} \\ \hline

  \end{tabular}
  \caption{Summary of simulation results (success rate, as \%). DD = Dense Descriptor. See Appendix for task success criteria and additional details.} 
  \label{table:simulation_results}
\end{table}

Table \ref{table:simulation_results} contains the results of the simulation experiments. 
%that in all of the simulated tasks 
Interestingly we find that our method's visual representation is capable of enabling policy learning that is remarkably close in performance to what can be achieved if the policy has access to ground truth world state information. 
In contrast the performances of the end-to-end (E2E) and autoencoder (AE) methods vary much more across the different tasks. 
%Sometimes E2E performs better, other times AE does.  
Since our method benefits from object mask information during visual training, we also experimented with letting the autoencoder use this information by applying the reconstruction loss on only the masked image.  Additionally we tried training the autoencoder end-to-end during behavior cloning.  Both of these yield mixed results, depending on the task.

Since the vision network in our method is a 34-layer ResNet, we wanted to see if the end-to-end method would benefit from using the same, deeper vision backbone. The deeper network did not improve closed-loop performance (Table \ref{table:simulation_results}) although it did reduce behavior-cloning validation error. This suggests the advantage of our method comes from the correspondence training rather than the model capacity.

\begin{figure}
\centering
\hspace*{0.0cm}
  \includegraphics[keepaspectratio=true,width=0.48\textwidth]{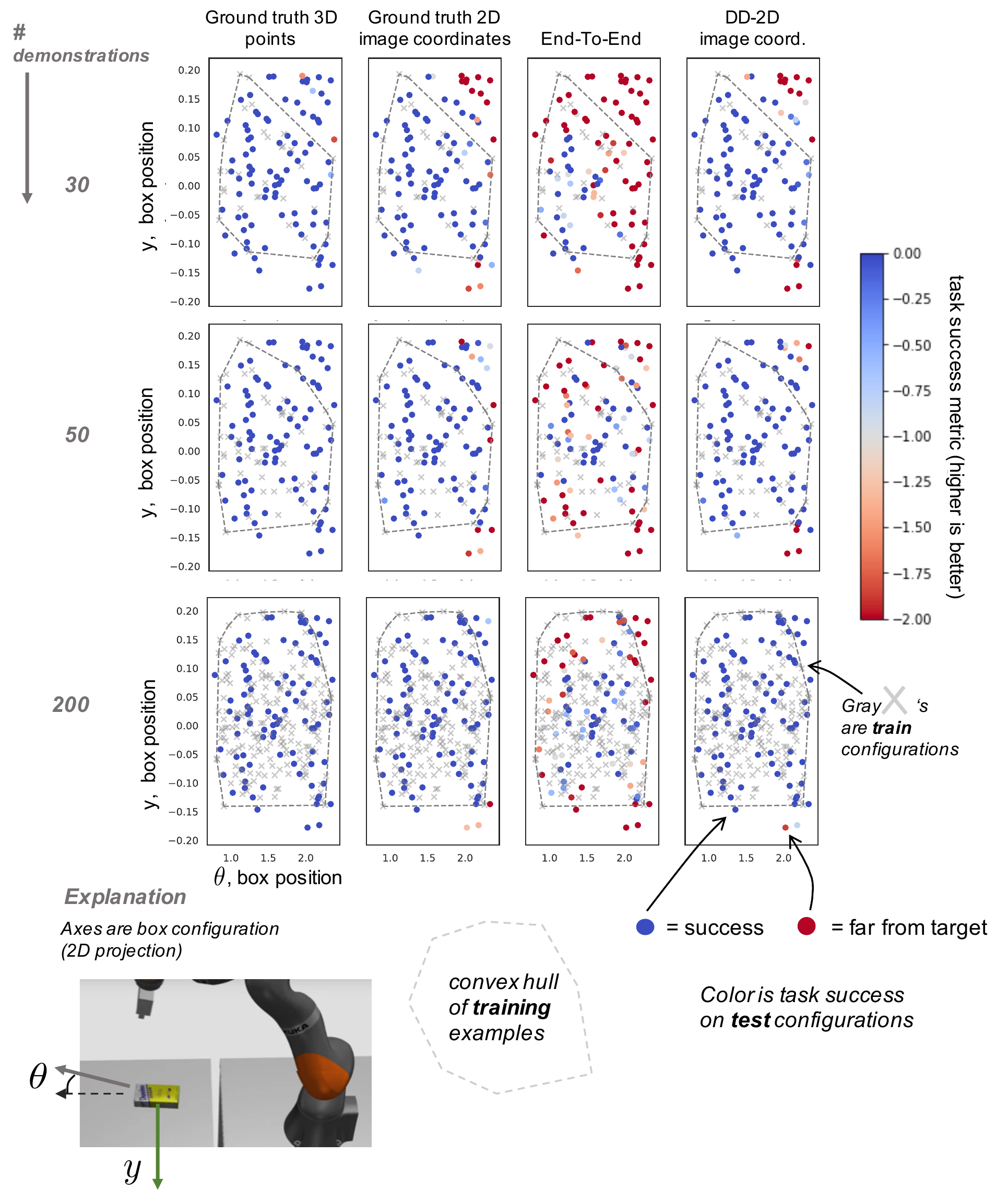}
  \captionof{figure}{Task success distribution plotted over the 2D projection of the varied box configurations for the \emph{``Reach, T + R''} task.  The color of each point
  represents the result of deploying the learned policy with the object at that $\theta, y$. Specifically the color encodes the distance to target threshold: $\min(0, -(\Delta \text{translation} + \Delta \text{rotation}) + \epsilon)$, where $\epsilon$ is the success threshold.
  %represents the result of deploying the learned policy with the reaching object target in the given $\theta, y$ position.  
  The $x$ coordinate is not shown in order to plot in 2D.  Dark blue corresponds to perfect performance on the task with the object in that configuration, red is poor performance.  Note that the color scale cuts off at -2 in order to highlight differences in the range [-2,0].  Each gray ``x'' in each subplot represents the configurations of the box in the training set, for either (from top to bottom) 30, 50, or 200 demonstrations.  The dashed gray line shows the convex hull of the respective training sets.}\label{MTBSE2spatial}
  \vspace{-0.3cm}
\end{figure}

The binary success metrics of Table \ref{table:simulation_results}, however, do not fully convey the methods' performances. We also experiment with varying the number of demonstrations, and characterize the performance distributions.  By plotting the performance for the \emph{``Reach, T + R''} task over a projection of the sampled object configurations (Figure \ref{MTBSE2spatial}), we learn that the few failures of our method occur when the box position lies outside the convex hull of the training data. Interestingly the GT-2D baseline also struggles with similar failure modes, while the GT-3D method succeeds in more cases outside the convex hull. This suggests that policies that consume 3D information are better able to extrapolate outside the training distribution; our DD-3D method also provides better generalization than DD-2D. The baseline vision-based methods do not generalize as well; for example, the E2E performance distribution is shown in Figure \ref{MTBSE2spatial}. On this task we find that with just 30 demonstrations our method outperforms both AE and E2E with 200 demonstrations.

The pushing tasks are of particular interest since they demand closed-loop visual feedback. Disturbances are applied to the object both while collecting demonstrations and deploying the learned policies. Since the \emph{``Push box''} task used a dynamic state feedback controller to provide demonstrations, we find that we need the sequence model (LSTM) for the policy network to achieve the task, even when the policy has access to ground truth object state.  On the other hand, the \emph{``Push plate''} task employed a static feedback controller to provide demonstrations, and so MLP models that consume only the current observation, $\pi_{\theta_p}(\bm{o}_t)$, are sufficient. 

Interestingly a variety of methods performed well on the \emph{``Push box''} task while large differences were evident in the \emph{``Push plate''} task. We speculate that this is because higher precision is required to accurately push the plate as compared to the box. Since the rectangular robot finger experiences a patch contact with the box, while only a point contact with the plate, there is more open loop stability in pushing the box.
On the harder \emph{``Push plate''} task we found that our DD-2D method performed almost as well as the GT-2D baseline and significantly outperformed both the AE and E2E approaches, and that DD-3D improved performance even further.

Additionally we find (Table~\ref{table:noise_results}) our noise augmentation technique (Sec.~\ref{subsec:trainig_for_feedback_through_data_augmentation}) has a marked effect on task success for behavior-cloned policies.  This applies to ground truth methods and our method, with as few as 30 demonstrations or as many as 200.

% NEW TABLE
\begin{table}
  \centering
  \footnotesize
  \begin{tabular}{|l|r|r|r|r|}
    \hline
    \multicolumn{1}{|l|}{} & \multicolumn{2}{l|}{\textit{30 demonstrations}} & \multicolumn{2}{l|}{\textit{200 demonstrations}} \\
    
    \multicolumn{1}{|l|}{Noise / Method} & \multicolumn{1}{l|}{GT-2D} & \multicolumn{1}{l|}{DD-2D} & \multicolumn{1}{l|}{GT-2D} & \multicolumn{1}{l|}{DD-2D}  \\ \hline \hline %\hhline{|=|=|=|=|=|}
    
    \textit{No noise}    & 5.6     & 1.1         & 5.6 & 3.4 \\ \hline
    \textit{With noise}  & \bf{73.3} & \bf{73.3} & \bf{95.6} & \bf{97.8} \\ \hline

  \end{tabular}
  \caption{Comparison of using our feedback-training noise augmentation technique or not (success rate, as \%) on the \emph{``Reach, T + R''} task. \textit{No noise} uses $\sigma_{i} = 0.0$, whereas \textit{With noise} uses $\sigma_{\text{translation}} = 1$mm, $\sigma_{\text{rotation}} = 1$ degree. See Section \ref{subsec:sim_experiments_and_benchmarking} for descriptions of GT-2D (ground truth) and DD-2D (our method).}
  \label{table:noise_results}
\end{table}

\subsection{Hardware Experimental Setup}
\label{subsec:real_hardware_experimentation}

%Our hardware closely resembles our simulation experimentation.  
We used a Kuka IIWA LBR robot with a Schunk WSG 50 parallel jaw gripper to perform imitation learning for the five tasks detailed in Figure \ref{hardwarefig}.  RGBD sensing was provided by RealSense D415 cameras rigidly mounted offboard the robot and calibrated to the robot's coordinate frame.  Note that for effective correspondence learning between views, it is ideal to have views with \textit{some} overlap such that correspondences exist, but still maintain different-enough views.  %Our hardware setup included three available cameras, %although in practice we would choose two of these cameras to log demonstration data for any given task.  Similar to simulation experimentation, we use 
%but for each task only two views were used for correspondence training, and policy training used only a single monocular RGB stream as image input to the policies. 
All shown hardware results use only RGB input for the trained policies (DD-2D, Sec.~\ref{subsec:sim_experiments_and_benchmarking}) and use descriptor set optimization (Sec.~\ref{subsec:correspondence-visuomotor}).
Human demonstrations were  provided by teleoperating the robot with a mouse and keyboard.

\subsection{Hardware Results}
\label{subsec:hardware_results}

\begin{table*}[t]
  \centering
  \footnotesize
  \resizebox{0.85\textwidth}{!}{
  \begin{tabular}{|l|c|c|c|c|c|c|c|c|c|c|}
\hline
         &                         &  Trained           & \multicolumn{3}{c|}{\footnotesize \textit{Without disturbances}}                  & \multicolumn{3}{c|}{\footnotesize \textit{With disturbances }} &     \multicolumn{2}{c|}{\footnotesize \textit{Demonstration data }}             \\ 
         &  Success         &   with manual                 &  \#                  &  \#                  &                                   &   \#                   &      \#               &                         & \# & time 	\\ 
Task  &  criterion         &   disturbances   &  attempts  &   success &  \%                             &  attempts  &   success &  \%                     & total & (min.)  \\ 
\hhline{|=|=|=|=|=|=|=|=|=|=|=|}

Push sugar & box is $<$ 3 cm   &  yes  & 6 &  6  &  100.0  &  70 & 68 & 97.1 & 51 & 13.9  \\  
box              & from finish line       &          &    &      &             &       &      &         & &  \\ \hline

Flip shoe,      & shoe is                 &  no    &  43  & 42 & 97.7 &  40 & 35 &  87.5 & 50 & 6.5 \\ 
single instance  & upright                &           &        &      &         &      &      &       & &   \\ \hline

\multicolumn{11}{|l|}{\footnotesize Flip shoe, } \\
\multicolumn{11}{|l|}{\footnotesize class-general } \\ \hdashline

{\textit {previously seen}}     & shoe is                 &  no    &  43  & 38 & 88.4 &  -- & -- &  --  & 146 & 17.5\\ 
{\textit {shoes (14)}}             & upright                 &           &        &      &         &      &      &      & &   \\ \hdashline

{\textit {novel}}            & shoe is                 &  no    &  22  & 17 & 77.3 &  -- & -- &  --  & 146 & 17.5\\ 
{\textit {shoes (12)}}    & upright                 &           &        &      &         &      &      &       & &  \\ \hline
                          
Pick-then-hang   & hat is                         &  yes    &  50  & 42 & 84.0 &  41 & 28 &  68.3 & 52 & 11.5 \\ 
hat on rack         & on the rack                 &           &        &      &         &      &      &   & &      \\ \hline
                           
Push-then-        & plate is                          &  yes    &  22  & 21 & 95.5 &  27 & 22 &  81.5  & 94 & 27.4\\ 
grab plate         & off the table                   &           &        &      &         &      &      &       & &  \\ \hline

{\textit{Total}}    &                                       &           & 186  &    &         & 178   &    &      & & \\ \hline
                          
% (49370 * 1 / 30.0) / 60.0 % to compute length in minutes
% 49730 * (640*480*3*8 + 13*32 + 6*32) / 8 % to compute bytes

  \end{tabular}}
  \caption{Summary of task attempts and success rates for hardware validation experiments.  Autonomous re-tries are counted as successes.} 
  \label{table_hardware_results}
\end{table*}

We validate both our visual learning method and its use in imitation learning in the real world.
%We validate that our learning approach works in the real world by performing extensive hardware experiments. %We experiment with our learning approach in the real world.  
As in simulation, we \emph{only use demonstration data} for both visual training and policy learning; no additional data collection is needed.
While the simulation results provide a controlled environment for comparisons, there are a number of additional challenges in our real world experiments: (i) visual complexity (textures, lighting, backgrounds, clutter), (ii) use of human demonstrations rather than expert simulation controllers, (iii) real physical contact, and (iv) imperfect correspondence learning due to noisy depth sensors and calibration.  Our hardware experiments test all of these aspects. All real hardware experiments use LSTM policy networks, since we suspect our human demonstrators use dynamic internal state.

\begin{figure}
\centering
\hspace*{-0.0cm}  
  \includegraphics[keepaspectratio=true,scale=0.19]{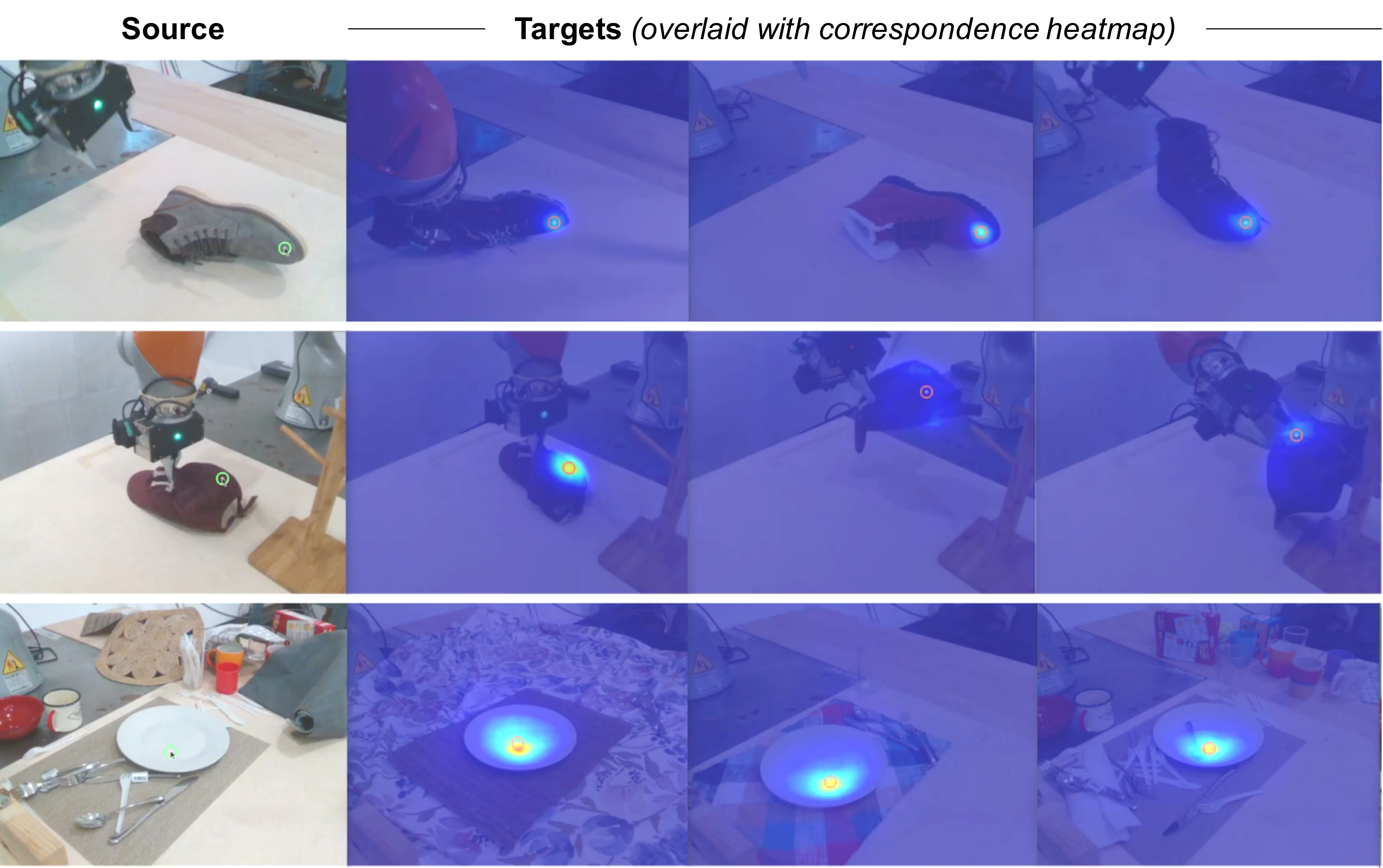}
  \captionof{figure}{Learned correspondences from demonstration data, depicted as correspondence heatmaps between a source pixel (left, with the green reticle)  and target scenes (right, with red reticle as best predicted correspondence).}\label{hardware_correspondences}
\end{figure}

\begin{figure}
\centering
\hspace*{-0.0cm}  
  \includegraphics[keepaspectratio=true,scale=0.17]{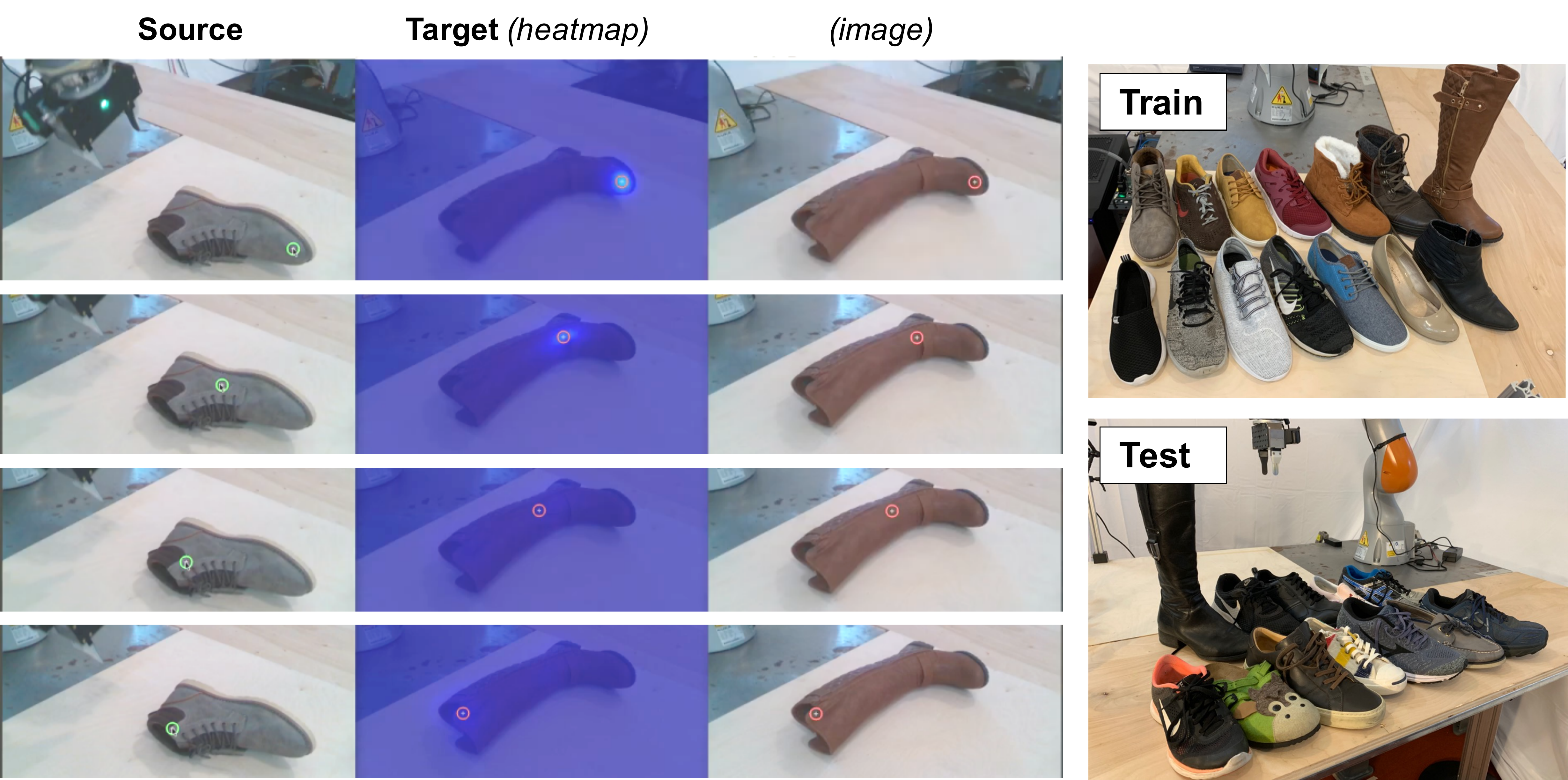}
  \captionof{figure}{Learned correspondences (left) between a standard-sized shoe and an extra-tall boot.  A small amount of movement near the top of the ankle on the shoe (far left) corresponds to a ``stretched-out'' movement on the boot (right).  Images cropped for visualization. Also shown are shoe train and test instances (far right).}\label{boot_correspondences}
\end{figure}

\subsubsection{Learned Correspondences from Dynamic Scenes}

Fig.~\ref{hardware_correspondences} displays visualizations of learned correspondences from demonstration data. The results show that the learned visual models, despite imperfect depth sensor noise, calibration, and only time-synchronized image pairs, are capable of identifying correspondences across a class of objects, for an object in different deformable configurations, and for objects in a diversity of backgrounds.   Figure \ref{boot_correspondences} displays class-general correspondences for a particularly challenging instance with large shape variation.%, and also displays our train and test sets for the class-general shoe task.

\subsubsection{Real-World Visuomotor Policies}

Figure \ref{hardwarefig} displays examples of autonomous hardware results, and Table \ref{table_hardware_results} provides a quantitative overview.  %Table \ref{table_hardware_overview} shows the slightly different observation and action space configurations used for the different hardware tasks, and also the number of demonstrations used, the cumulative total demonstration time, and the total size of the data used for policy training.
%Table \ref{table_hardware_results} provides a coarse quantitative overview of the results.  
To highlight a few results, several of the tasks achieve over 95\% reliability, including the \emph{``Push sugar box''} task with and without disturbances, and the \emph{``Flip shoe, single instance''} and \emph{``Push-then-grab plate''} tasks without disturbances. Each of the different tasks present significant challenges, best appreciated in \href{https://sites.google.com/view/visuomotor-correspondence}{our video}. Several of the tasks include non-prehensile manipulation, including pushing the box and plate, and flipping the shoes.  In the \emph{``Pick-then-hang hat on rack''} task, the robot autonomously reacts to the deformable configuration of the hat after disturbances.  %Although the reliability is not as high as other tasks, this is perhaps the most challenging of the tasks given the current state of the art in robotics.  
The \emph{``Push-then-grab plate''} task as well is highly challenging given the visual clutter, the symmetry and lack of visual texture for the object, and requires using ``extrinsic dexterity'' \cite{dafle2014extrinsic} via the wood block to enable sliding the gripper into position to grasp the plate.

\section{CONCLUSION}

Our experiments have shown self-supervised correspondence training to enable efficient policy learning in the real world, and our simulated imitation learning comparisons empirically suggest that our method outperforms two vision-based baselines in terms of generalization and sample complexity.  While different hyperparameters, model architectures, and other changes to the baselines may increase their performance, our method is already near the upper bound of what can be expected in the used experimental setting: it achieves results comparable to baselines using ground truth information.  One reason our approach may outperform the vision-based baselines is that it additionally uses a fundamentally different source of supervision, provided by visual correspondence training.  Since our approach is self-supervised, it does not entail additional human supervision.  

Dense descriptor learning has shown to be an exciting route for improving visuomotor policy learning.  While this has enabled the variety of tasks shown, there are many that are out of scope.
One current limitation is that our visual representation does not explicitly address simultaneously viewing multiple object instances of the same class.  Future work could, similar to the visual pipeline in \cite{manuelli2019kpam}, combine both instance-level segmentation with intra-instance visual representations.  Additionally, returning to the cooking eggs example in the Introduction, it is interesting to consider using spatial correspondence as part, but not the entirety, of the visual representation of the world.


%%%%%%%%%%%%%%%%%%%%%%%%%%%%%%%%%%%%%%%%%%%%%%%%%%%%%%%%%%%%%%%%%%%%%%%%%%%%%%%%

%%%%%%%%%%%%%%%%%%%%%%%%%%%%%%%%%%%%%%%%%%%%%%%%%%%%%%%%%%%%%%%%%%%%%%%%%%%%%%%%


% It will be hard to have an appendix also fit in 6 pages.

\section*{APPENDIX}

%\subsection{Evaluating Visual Representations for Visuomotor Policy Learning}

%\noindent \textbf{Metrics for Evaluating Visual Representations for Visuomotor Policy Learning}
\textbf{Simulation Tasks.}
%In simulation we performed four tasks. 
Our simulation environment was configured to closely match our real hardware experimentation.  Using Drake \cite{drake}, we simulate the 7-DOF robot arm, gripper, objects, and multi-view RGBD sensing. %with two cameras as shown in Fig.~\ref{fig:sim_experiments_images}. 
\emph{\textbf{``Reach T only''}}: goal is to move the end-effector to a target position relative to the sugar box object; success is within 1.2cm of target. The box pose only varies in translation, not rotation; training positions drawn from a truncated Gaussian ($\sigma_x=5$cm,$\sigma_y=10$cm), centered on the table, truncated to a 40cm$\times$10cm region. Test distribution drawn from uniform over same region. \emph{\textbf{``Reach T + R''}}: same as \emph{``Reach T only''} but now the box pose varies in rotation as well, drawn from a uniform [-30,30] degrees; success is within 1.2cm and 2 degrees. \emph{\textbf{``Push box''}}: goal is to push the box object across the table, and the box is subject to random external disturbances; success if translated across table and final box orientation is within 2 degrees of target. \emph{\textbf{``Push plate''}}: goal is to push a plate across a table to a specific goal location, and the plate is subject to external disturbances; success if plate center is within 1cm of target position.

\textbf{Policy Networks}:
All experiments using an \emph{\textbf{``MLP"}} had a two-layer network with 128 hidden units, 20\% dropout, in each layer and ReLU nonlinearities. Training was 75,000 steps with RMSProp, $\alpha=0.9$, with a batch size of 16, and \textit{lr} starting at $1e-4$, and decaying by a factor of 0.5 every 10,000 steps. 
All experiments using an \emph{\textbf{``LSTM"}} had a single LSTM layer with 100 units preprocessed by two MLP layers of 100 units, 10\% dropout, and layer-normalized prior to the LSTM layer. Training was 200,000 steps with RMSProp, $\alpha=0.9$, with \textit{lr} starting at $2e-3$, decaying 0.75 every 40,000 steps, with truncated backpropagation of maximum 50 steps, and gradient clipping of maximum magnitude 1.0. As recommended in \cite{rahmatizadeh2018vision} we train LSTMs on downsampled trajectories, we use 5 Hz.

\textbf{Vision Networks}:
Both \emph{\textbf{``AE''}} and \emph{\textbf{``E2E''}} methods used an identical architecture, with the only difference being the additional decoder used for the AE method during autoencoding. The network is almost exactly as in \cite{levine2016end} and \cite{finn2016deep}, but we provided a full-width image, $320\times240$. We used 16 2D feature points.  %($240\times240$ gives comparable results, but we report the full-width results to better compare with our method which uses full-resolution $640\times480$.)
\emph{\textbf{``DD''}} architecture is identical to \cite{florence2018dense}. DD-2D computes image-space spatial expectation, DD-3D computes 3D-space spatial expectation using the depth image, see \cite{florence2019thesis} for details; both used 16 descriptors.  The \emph{\textbf{``E2E (34-layer)''}} network is exactly the DD architecture but with $D=16$ and channel-wise 2D spatial softmax to obtain $\bm{z}$.

\section*{ACKNOWLEDGMENT}

This work was supported by National Science Foundation Award No. IIS-1427050, Lockheed Martin Corporation Award No. RPP2016-002, and an Amazon Research Award grant. The views expressed are not endorsed by our funding sponsors.

%%%%%%%%%%%%%%%%%%%%%%%%%%%%%%%%%%%%%%%%%%%%%%%%%%%%%%%%%%%%%%%%%%%%%%%%%%%%%%%%

% {\small
% \bibliographystyle{ieee}
% \bibliography{all-bib.bib}
% }

\bibliographystyle{IEEEtran}
\bibliography{IEEEabrv,all-bib}

\end{document}